\documentclass[10pt,twocolumn,letterpaper]{article}

\usepackage[pagenumbers]{cvpr}

\usepackage[dvipsnames]{xcolor}

\usepackage{epsfig}
\usepackage{graphicx}
\usepackage{comment}
\usepackage{amsmath,amssymb} %
\usepackage{xcolor}
\usepackage[export]{adjustbox}
\usepackage{bbm}

\usepackage[font=small,skip=5pt]{caption}
\usepackage{float}

\usepackage{multirow}
\usepackage{tabularx}
\usepackage{soul}
\usepackage{array}
\usepackage{comment}
\usepackage{outlines}

\usepackage[mathscr]{euscript}
\newcolumntype{L}[1]{>{\raggedright\let\newline\\arraybackslash\hspace{0pt}}m{#1}}
\newcolumntype{C}[1]{>{\centering\let\newline\\arraybackslash\hspace{0pt}}m{#1}}
\newcolumntype{R}[1]{>{\raggedleft\let\newline\\arraybackslash\hspace{0pt}}m{#1}}
\newcolumntype{Y}{>{\centering\arraybackslash}X}
\usepackage{booktabs}
\usepackage{enumitem}

\usepackage{fancyhdr}
\usepackage{overpic}
\usepackage{wrapfig}

\usepackage{soul}
\definecolor{cb-black}      {RGB}{  0,   0,   0}
\definecolor{cb-blue-green} {RGB}{  0,  073,  073}
\definecolor{cb-green-sea}  {RGB}{  0, 146, 146}
\definecolor{cb-rose}       {RGB}{255, 109, 182}
\definecolor{cb-salmon-pink}{RGB}{255, 182, 119}
\definecolor{cb-purple}     {RGB}{ 73,   0, 146}
\definecolor{cb-blue}       {RGB}{ 0, 109, 219}
\definecolor{cb-lilac}      {RGB}{182, 109, 255}
\definecolor{cb-blue-sky}   {RGB}{109, 182, 255}
\definecolor{cb-blue-light} {RGB}{182, 219, 255}
\definecolor{cb-burgundy}   {RGB}{146,   0,   0}
\definecolor{cb-brown}      {RGB}{146,  73,   0}
\definecolor{cb-clay}       {RGB}{219, 209,   0}
\definecolor{cb-green-lime} {RGB}{ 36, 255,  36}
\definecolor{cb-yellow}     {RGB}{255, 255, 109}

\usepackage{amsmath}
\usepackage{siunitx}

\newcommand{\refg}[1]{\textcolor{red}{{#1}}}

\newcommand{\PAR}[1]{\vskip4pt \noindent{\bf #1~}}
\newcommand{\PARnospace}[1]{\vskip4pt \noindent{\bf #1}}
\newcommand{\ours}{iReplica}
\newcommand{\oursbig}{Interaction Replica (iReplica)}

\usepackage{amsthm}

\usepackage{graphicx}
\usepackage{amsmath}
\usepackage{amssymb}
\usepackage{booktabs}
\usepackage{float}

\definecolor{cvprblue}{rgb}{0.21,0.49,0.74}
\usepackage[pagebackref,breaklinks,colorlinks,citecolor=cvprblue]{hyperref}

\def\paperID{68} %
\def\confName{3DV\xspace}
\def\confYear{2024\xspace}

\newcommand{\myvec}[1]{\mathbf{#1}}

\title{\vspace{-5mm}Interaction Replica: Tracking Human–Object Interaction and Scene Changes From Human Motion\vspace{-5mm}}

\author{\begin{tabular}{cccccccccccc}\multicolumn{3}{c}{Vladimir Guzov \textsuperscript{1,2}} & \multicolumn{3}{c}{Julian Chibane \textsuperscript{1,2}} & \multicolumn{3}{c}{Riccardo Marin \textsuperscript{1}} & \multicolumn{3}{c}{Yannan He \textsuperscript{1}} \\  \multicolumn{4}{c}{\hspace*{8.5mm}Yunus Saracoglu \textsuperscript{1}} & \multicolumn{4}{c}{\hspace*{4.5mm}Torsten Sattler \textsuperscript{3}} & \multicolumn{4}{c}{Gerard Pons-Moll\textsuperscript{1,2}} \end{tabular}\\\\
{\small \textsuperscript{1}Tübingen AI Center, University of Tübingen, Germany, \qquad \textsuperscript{2}Max Planck Institute for Informatics, Saarland Informatics Campus, Germany}  \\
{\small\textsuperscript{3}CIIRC, Czech Technical University in Prague, Czech Republic}\\
{\tt\scriptsize \{vladimir.guzov, riccardo.marin, yannan.he, gerard.pons-moll\}@uni-tuebingen.de,} \vspace{-2mm}\\ {\tt\scriptsize jchibane@mpi-inf.mpg.de, yunus.saracoglu@student.uni-tuebingen.de, torsten.sattler@cvut.cz}}

\begin{document}
\twocolumn[{%
\renewcommand\twocolumn[1][]{#1}%
\maketitle
\begin{center}
    \centering
    \captionsetup{type=figure}
    \vspace{-30pt}
    \includegraphics[width=0.94\textwidth]{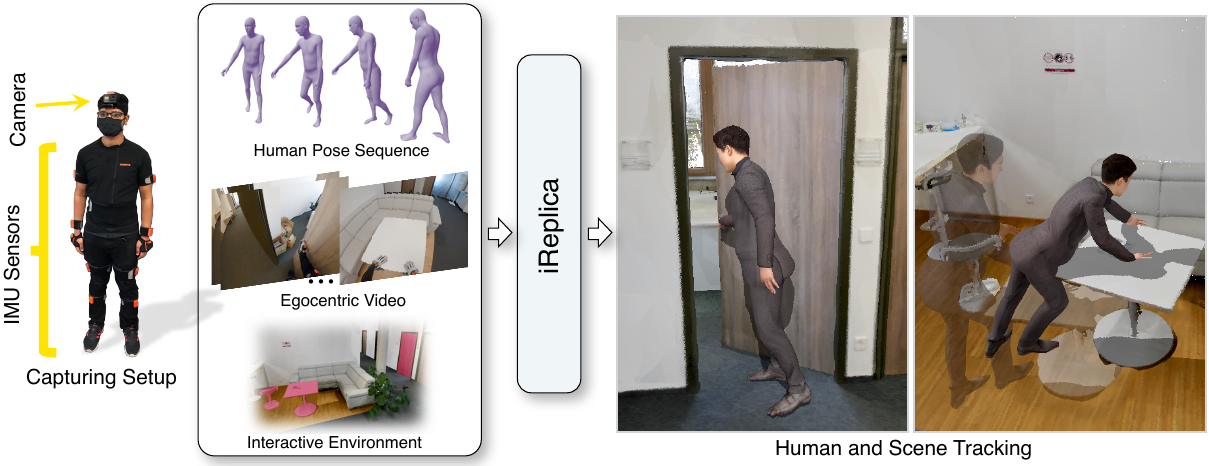}
    \caption{\textbf{\oursbig{}}. iReplica estimates location and full 3d pose of a subject within a large 3D scene and dynamically tracks changes made to the scene by the subject - using only wearable sensors (\textit{left}), removing the need of external sensors.
    We obtain an approximate 3D human pose sequence using IMU sensors and use head camera self-localization to localize the subject in the prescanned 3d interactive environment scene.
    iReplica predicts human-scene contacts and updates the scene in case of interaction.}

    \label{fig:teaser}
\end{center}%
}]

\thispagestyle{empty}

\begin{abstract}
\vspace{-5mm}
Our world is not static and humans naturally cause changes in their environments through interactions, e.g., opening doors or moving furniture. Modeling changes caused by humans is essential for building digital twins, e.g., in the context of shared physical-virtual spaces (metaverses) and robotics. In order for widespread adoption of such emerging applications, the sensor setup used to capture the interactions needs to be inexpensive and easy-to-use for non-expert users. I.e., interactions should be captured and modeled by simple ego-centric sensors such as a combination of cameras and IMU sensors, not relying on any external cameras or object trackers. Yet, to the best of our knowledge, no work tackling the challenging problem of modeling human-scene interactions via such an ego-centric sensor setup exists. This paper closes this gap in the literature by developing a novel approach that combines visual localization of humans in the scene with contact-based reasoning about human-scene interactions from IMU data. Interestingly, we can show that even without visual observations of the interactions, human-scene contacts and interactions can be realistically predicted from human pose sequences. Our method, iReplica (Interaction Replica), is an essential first step towards the egocentric capture of human interactions and modeling of dynamic scenes, which is required for future AR/VR applications in immersive virtual universes and for training machines to behave like humans. Our code, data and model are available on our project page at \href{http://virtualhumans.mpi-inf.mpg.de/ireplica/}{http://virtualhumans.mpi-inf.mpg.de/ireplica/}.
\end{abstract}

\vspace{-0.7cm}
\section{Introduction}
\label{sec:intro}
Current augmented and virtual reality (AR/VR) applications show promising potential: interesting applications include collaborative developments, virtual meeting rooms, and personal assistants that help users navigate the world. While it is clear that for an immersive experience blending real and digital worlds is crucial, the current AR/VR experience is restricted to small spaces, i.e., in general, a few square meters, possibly free from objects. But consider daily actions like moving across rooms, opening and closing doors, or gathering chairs around a table. Even these simple actions are not easy to capture with present technology, which limits the scope of AR/VR applications.

\begin{figure}[ht!]
    \centering
    \begin{overpic}[trim=0cm 0cm 0cm 0cm,clip, width=0.88\linewidth]{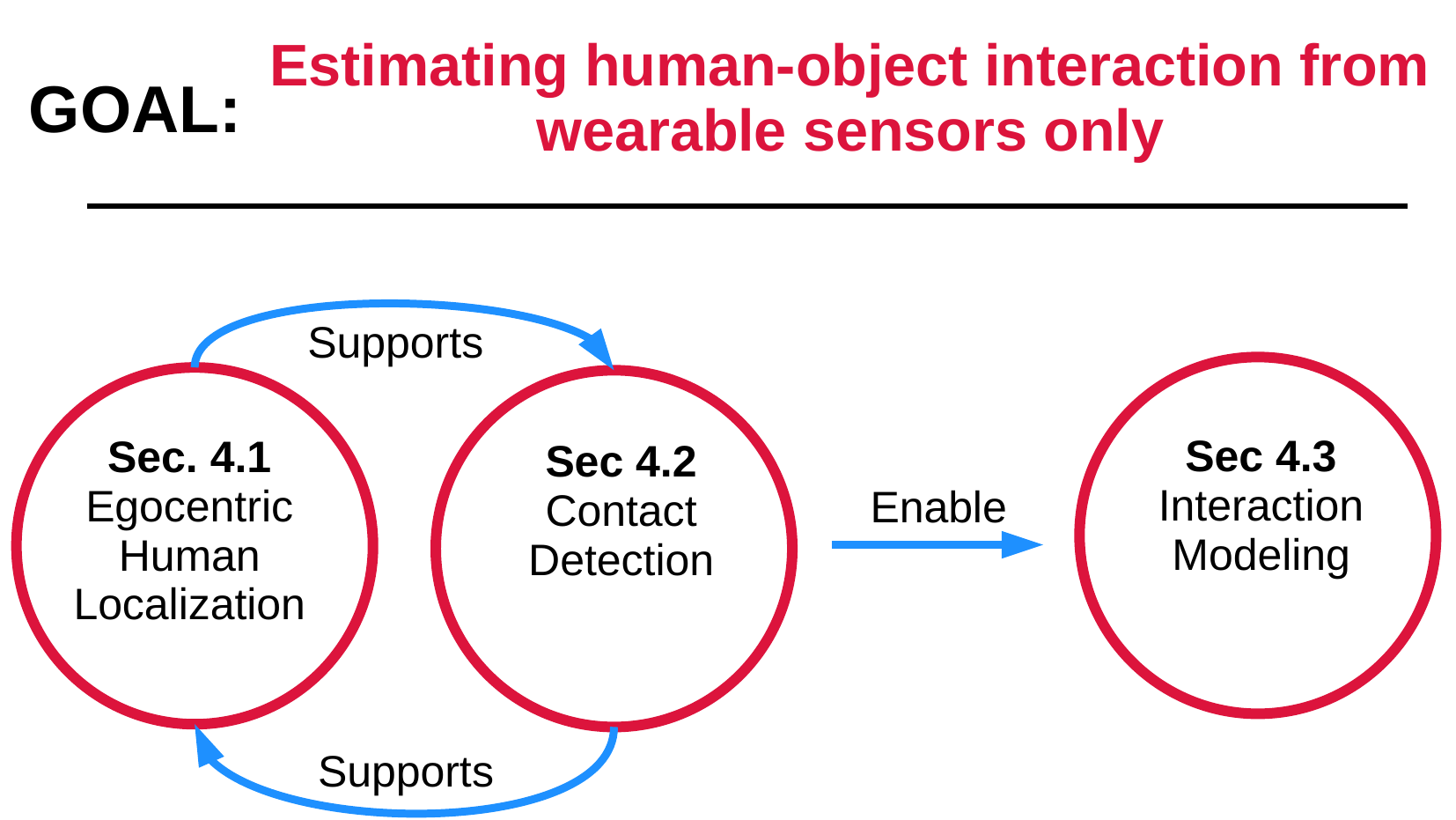}
    
    \end{overpic}
    
    \caption{\label{fig:subdiv} \textbf{Problem subdivision.} We demonstrate that joint integration of different sub-research problem improves and support each other. We show this is fundamental to achieve our goal of estimating human-scene interaction from wearable sensors only.}
\end{figure}

The predominant approach for 3D human motion estimation relies on external cameras~\cite{kanazawa2018endtoend, joo2018total, pavlakos2018humanshape, guler2018densepose, iskakov2019learnable, kocabas2020vibe, zhang2021learning, luo2022embodied}. Yet, asking non-technical users to mount and calibrate complex multi-camera systems is clearly infeasible. Body-mounted sensors, \eg, cameras and IMU sensors, seem much more ready for mass adoption.

Prior ego-centric trackers such as HPS~\cite{guzov2021human}, EgoLocate~\cite{EgoLocate2023} or HSC4D~\cite{dai2022hsc4d} estimate human movement and position the person by combining head camera visual localization with IMU-based pose estimation. Methods like HPS, however, do not track scene changes. For example, if a person opens and walks through a door, such methods will only localize the person but can not infer the door movement, creating implausible reconstructions; see Figure~\ref{fig:qualitative_comparison}, HPS. 

\begin{figure*}[t!]
\centering
    \begin{overpic}
    [trim=0cm 0cm 0cm 0cm,clip,width=0.95\linewidth]{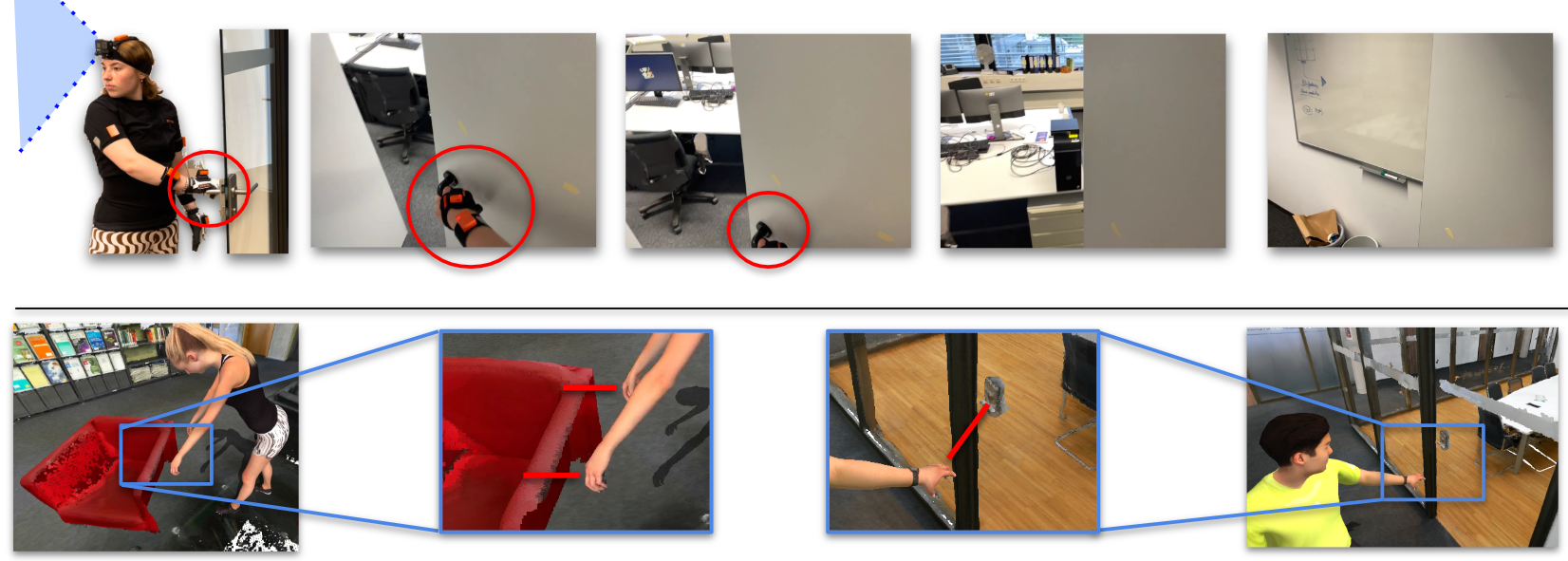}
    \put(27.5, 0){\small{Interaction is visually implausible due to localization offset}}
    \put(1.5, 17){\small{Interaction is out of view}}
    \put(28, 17){\small{Interaction is visible only partially}}
    \put(70, 17){\small{Interaction is out of view}}
    \put(5, 35){\small{Schema}}
    \put(20, 35){\small{Egocentric camera view}}
    \end{overpic}
    
    \caption{\label{fig:challenges} \textbf{Challenges.} Top row: The prediction of human-scene contacts (red circles) is hard because the interactions are frequently not in the camera view.
    Bottom row: Virtual replica of human pose and localization by prior work HPS~\cite{guzov2021human}. HPS achieved great progress in localizing humans solely by wearable sensors (camera+IMUs). However, for our task, the localization error of 4--16~cm (red lines) leads to visually implausible results for scene interactions.}
    \vspace{-3mm}
    
\end{figure*}

In this work, we address, for the first time, the problem of human-scene interaction capture \emph{from wearable sensors only}. Localizing the person with sufficient precision to track scene changes is hard, let alone estimating object motion. A major challenge is that the object is often not visible or is only partially visible in the camera; see Figure~\ref{fig:challenges}. In addition, since the head camera is in motion, the object's motion relative to the static world can not be directly inferred. 

\emph{Since no external sensors can measure the scene changes directly, how can we predict them?} 
Our key observations and findings are that 1) contact poses are distinctive and can be detected without visual clues, 2) knowing contact time stamps can regularize human localization, 3) objects move when the human contacts them\footnote{In this work we only consider static objects moved by the captured human.}. Motivated by this, we propose \emph{iReplica - Interaction Replica}, a novel human-centric method that automatically localizes the human in the scene (1. egocentric human visual localization), detects the time of contact and release with the object (2. contact time detection), and infers object motion based on contacts and human motion (3. interaction modeling). While works exist in each of these three sub-areas of research, no work integrates them simultaneously.

We needed several scientific innovations to integrate the aforementioned three sub-areas of research successfully. First, we improve human visual localization (HPS~\cite{guzov2021human}) by optimizing the human trajectory to match reliable head camera poses and detect spatio-temporal contacts. Second, we train a transformer-based contact time detection approach based solely on the human pose, which achieves a remarkable accuracy of 0.91 and an average precision of 0.81. 
Third, based on the refined human visual localization in the scene and the accurate contact predictions, we infer object motion coherent with the human. Our results demonstrate that joint integration is beneficial (Fig.~\ref{fig:subdiv}). The contact time information can be used to regularize visual localization by forcing the virtual human to contact the scene. Having precise human localization in the scene, along with contact timestamps, allows us to infer 1) where contacts occur and 2) the object's motion without seeing the object or contacts in the camera.

During this project, we captured two new datasets. To train a contact detection method, we captured a dataset of 8 subjects and more than $3$ hours of human-scene interactions annotated with contact time stamps. To validate our proposed method, we captured a dataset with subjects moving and interacting with different objects in large 3D scenes.
Our experiments show that \ours{} can capture, for the first time, full interactions, including the human motion, its location within the 3D scene and the scene changes,  
all from wearable sensors alone. We demonstrate that our human-centric approach outperforms baselines, which rely on SOTA camera-based contact detection or visual object localization~\cite{VISOR2022,shan2020understanding}.

In summary, our contributions are the following:
\begin{enumerate}
\item \textit{Novel Problem \& Method}: We are the first to tackle capturing human-scene interactions while localizing the human in the scene from wearable sensors alone. We propose a method to address this problem, obtaining, \textit{for the first time}, a digital replica of the human interaction in the scene without any external cameras.

\item \textit{Novel Data \& Metrics}: We provide H-contact -- a dataset of 2300+ human-scene interactions with ground truth annotated contacts. Additionally, we provide EgoHOI -- a dataset of human-scene interactions in scanned environments. We propose metrics to measure the visual plausibility of reconstructed interactions and the accuracy of contact prediction and object localization.

\end{enumerate}

To foster progress in this new research area, we release the method code, the evaluation protocol, and the datasets, including scans of the scenes, human motion capture aligned with them, and annotated contact timestamps. 

\section{Related Work}
\label{sec:related_work}

\PAR{Human--object interaction.}
Majority of current methods that record human--object interactions use external cameras. 
Methods to capture the full body pose also use external cameras and mostly static scenes~\cite{PROX:2019, zhang2021learning, luo2022embodied, huang2022capturing} or work with a single dynamic object without any scene context~\cite{zhang2020phosa, xu2021d3dhoi, weng2021holistic, taheri2020grab, huang2022intercap, bhatnagar2022behave}; similarly for human--scene and human--object interaction generation methods~\cite{taheri2022goal, zhang2022couch, hassan2021stochastic, starke2019neural}. A few exceptions exist though, \eg RigidFusion \cite{wong2021rigidfusion} can track objects with an external RGBD sensor, and 
some pose estimation methods work with first-person view footage.
However, those methods study the upper body or are limited to static cameras, \eg, hand--object pose estimation~\cite{tekin2019h+, kwon2021h2o, liu2021semi, doosti2020hope, oberweger2019generalized, wang2019learning}, or do not model dynamic objects~\cite{zheng2022gimo}. Some methods can use egocentric video to predict the contact ~\cite{VISOR2022,shan2020understanding}, but they do not further process this information to infer object position.
Our method works with body-mounted sensors and a 
moving camera while capturing the full-body pose and object position. 

\PAR{Embodied research.} 
Body-mounted sensor setups are heavily used to solve various tasks:
activity recognition methods
~\cite{ijcai2017-200, fathi2011understanding,ma2016going,cao2017egocentric,yonemoto2015egocentric,rogez2015first} use egocentric cameras looking at the body. 
However, they typically concentrate on capturing the upper body. Many full-body capturing methods~\cite{rhodin2016egocap,xu2019mo2cap2,SelfPose2020} work with similar head-mounted setups, but as the cameras are pointed at the person wearing them rather than outwards,
these methods ignore the environment around the subject.
Some methods work with an outwards-facing camera~\cite{jiang2017seeing,yuan20183d,Yuan_2019_ICCV}. 
However, they do not use any additional sensors to capture the body pose and predict it using motion priors. 
On the opposite side, methods like~\cite{yi2021transpose, jiang2022transformer} use inertial sensors to capture the body pose, but lack of visual cues results in an accumulating position drift, and body poses far from the ground truth. \cite{liu2022egocentric} proposes action recognition and localization using a first-person perspective video but does not model scene change. \cite{zhang2022egobody} works with a head-mounted camera, but uses it to capture the pose of other subjects, making it an external-camera method.
Most related to our method, HPS~\cite{guzov2021human} and following works~\cite{EgoLocate2023, dai2022hsc4d} capture human motion using body-mounted IMUs and a head-mounted camera looking outwards to capture the subject location within a pre-built 3D scan of the environment. 
We extend HPS by not only tracking the human pose but also an object the person interacts with.
Whereas HPS is restricted to static environments and cannot model scene changes caused by human--object interactions, \ours{} removes these restrictions.

\vspace{-0.1cm}
\PAR{Visual localization.} 
Visual localization aims to estimate the pose of a camera in a known environment. 
Current state-of-the-art approaches are based on 2D--3D matches between pixels in the camera image and 3D scene points. 
These 2D--3D matches are either estimated based on local features~\cite{sarlin2019coarse,Sattler2017PAMI,Schoenberger2018CVPR,Irschara09CVPR,Li2012ECCV,Lynen2020IJRR} or by regressing a 3D point coordinate for each pixel~\cite{Shotton2013CVPR,brachmann2020ARXIV,Brachmann2018CVPR,Cavallari2019TPAMI,dong2021robust}\footnote{We refer the interested reader to~\cite{Brachmann2021ICCV} for a discussion and comparison of both types of approaches.}. 
A recent line of works focuses on the robustness of localization algorithms~\cite{Toft2020TPAMI,wald2020beyond,Jafarzadeh2021ICCV,dong2021robust}, \eg, 
to illumination, weather, and seasonal changes, as well as to changes caused by human actions (rearranging furniture, \etc). 
These approaches assume that a large enough part of the scene remains static and is observed by the camera to facilitate pose estimation. 
The second assumption is violated in our scenario and we use IMU-based human pose tracking to bridge gaps where visual localization algorithms will most likely fail. 
As in~\cite{guzov2021human}, for the head-mounted camera localization, we use a state-of-the-art localization pipeline~\cite{sarlin2019coarse,sarlin2020superglue}. Similar to the idea of visual-inertial approaches, \eg ~\cite{Lynen2015RSS,Jones2011IJRR, ORBSLAM3_TRO}, we use data from the IMU sensor to stabilize the predicted camera trajectory during periods of low scene visibility or when a lot a scene changes are happening.
Note that our main contribution, \ie, jointly reasoning about human and object pose, is not tied to any particular localization algorithm.

\label{sec:method}

\section{Problem Setting}
\paragraph{Goal.} We aim to \emph{estimate human-object interaction from wearable sensors only}, without information from external sensors, using only body-mounted IMUs and an egocentric camera.
This opens a broad set of interconnected challenges: how do we define the interaction? How do we detect the start and the end of it? And how do we track the object's motion without having sensors dedicated? 

\PAR{Assumptions.} We assume a static 3D scan of the scene, along with a set of marked interactive objects, knowing their initial position and degrees of freedom (\eg, a sofa can slide on the ground but cannot be lifted, or a door rotates around a hinge). We refer to this as \emph{interactive environment} (IE).

\PAR{Input/Output.} We require a set of body-mounted wearable IMUs (we use 17 sensors from XSens~\cite{xsensawinda}) and a video stream from a head-mounted camera. 
Relying only on wearable sensors lets us handle large scenes consisting of multiple rooms. 
Compared to external cameras, wearable sensors are much more consumer-friendly as they are easier to set up.
\ours{} outputs a virtual replica of the interaction, \ie, coherent human and object motion in the scene.

\section{iReplica} 
\label{sec:ireplica}
\begin{figure*}[h!]
\vspace{-2mm}
\small
    \centering
    \begin{overpic}[trim=0cm 0cm 0cm 0cm,clip, width=0.95\linewidth]{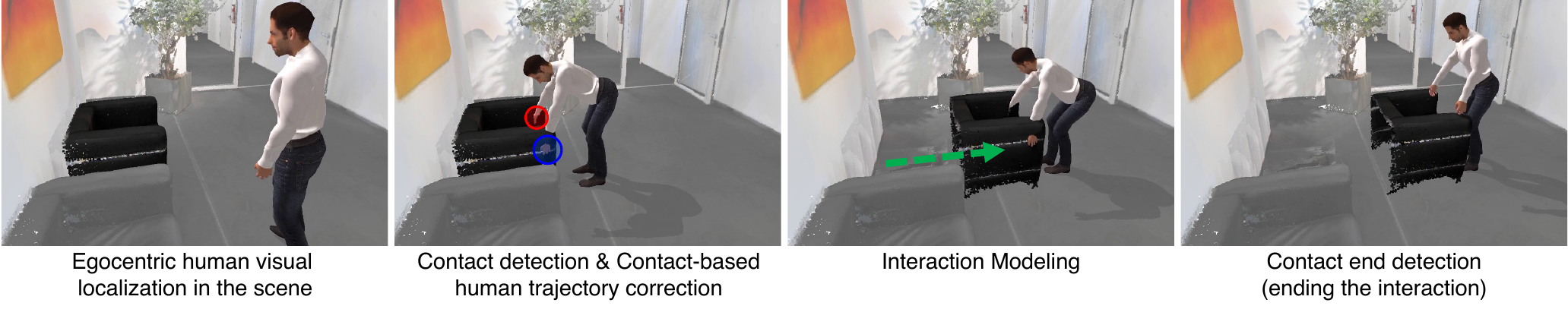}
    \put(11,21.5){A)}
    \put(36.5,21.5){B)}
    \put(62,21.5){C)}
    \put(87,21.5){D)}
	\end{overpic}

    \vspace{-3.5mm}
    \caption{\textbf{Overview of \ours{}.} \ours{} estimates a subject's location and full pose within a large 3D scene and dynamically track changes made to the scene by the subject -- using only wearable sensors.  We do so in 4 steps: \textbf{A)} We obtain an initial localization of the subject in the IE by head camera self-localization. \textbf{B)} The start of the interaction is predicted by a neural network. Predictions are provided as contact / no-contact classification of the subject's hands (red and blue areas). The contacts are used to correct head camera localization of the subject, snapping the human trajectory smoothly to the object. \textbf{C)} The motion of the contacted regions is used to infer the object trajectory (green). \textbf{D)} The network predicts the release, essential to stop object dragging. The algorithm is detailed in Sec.~\ref{sec:ireplica}.}
    \label{fig:overview}

    \vspace{-0.5cm}
\end{figure*}

\PAR{Overview.} 
We obtain initial localization and pose estimation for the person relying on an improved version of HPS~\cite{guzov2021human} (Sec~\ref{sec:localization}, Fig~\ref{fig:overview} A). Our method considers only the human pose at each instant and predicts the probability of contact with an object (Sec~\ref{sec:contact}, Fig~\ref{fig:overview} B). Once the contact is detected, we model the object dynamically as follows: we deform the human trajectory to match the object contact; the object is attached to the human and driven in space according to its degrees of freedom (Sec~\ref{sec:modeling}, Fig~\ref{fig:overview} C); when our method infers from the human pose the end of the contact, the object is released (Fig~\ref{fig:overview} D). 

\subsection{Egocentric human visual localization.} 
\label{sec:localization}

\PAR{Problem.} Our method is built on a combination of IMUs and head-mounted camera data. Previous methods rely on optimizations to get from these two modalities smooth trajectories estimation~\cite{guzov2021human,Jones2011IJRR,Lynen2015RSS}. However, no previous approach considers the human's interaction with the scene nor shows extensions to incorporate constraints coming from this. Also, if 10 cm of error (average for HPS~\cite{guzov2021human}) might not seem much for human localization in a building, for human-object interaction (which is our ultimate goal), this can cause dramatic inconsistencies. Instead, we see (and take advantage of) the relation between human localization and contact prediction: solving for contact prediction supports human localization in large volumes; human localization helps detect object contact in time and space. 

\PAR{Trajectory optimization.} We start introducing an improvement over the HPS approach~\cite{guzov2021human}. We deploy a simple optimization that is flexible and can be used to incorporate interaction constraints. While we work with 3D trajectories, we consider a 2D optimization since one dimension (gravity axis) is constrained by the ground of the scene. 
Consider the trajectory described as a 2D curve $\myvec{l}(\tau)=(x(\tau), y(\tau))$ defined in the time interval $ \tau=[\tau_\mathrm{start},\tau_\mathrm{end}]$, and a list of $K$ control points 
${\myvec{p}}=\{{\myvec{p}}_i=({x}_i, {y}_i)\}_{i=1}^K$ (constraints) at times
$\tau_\mathrm{start} \leq {\tau}_1,\hdots,{\tau}_K \leq \tau_\mathrm{end}$. We want to recover a new trajectory $\hat{\myvec{l}}(\tau)=(\hat{x}(\tau), \hat{y}(\tau))$ that gets close to the control points while not deviating too much from the initial trajectory.
We introduce an energy $E_{tr}$ that encodes the trajectory deviation in terms of angles.
\begin{equation}
    E_{tr}(\hat{\myvec{l}},\myvec{l})= \int_{\tau_\mathrm{start}}^{\tau_\mathrm{end}} \left(\frac{d\hat{\alpha}(\tau)}{d\tau}-\frac{d\alpha(\tau)}{d\tau}\right)d\tau\space,
    \label{eq:bending}
\end{equation}
where:
\begin{equation*}
\hat{\alpha}(\tau) = \operatorname{atan2}\left(\frac{d\hat{y}}{d\tau},\frac{d\hat{x}}{d\tau}\right), \quad \alpha(\tau) = \operatorname{atan2}\left(\frac{dy}{d\tau},\frac{dx}{d\tau}\right).
\end{equation*}
Concretely, $E_{tr}$ measures the difference between two trajectories at each instant in terms of direction (angle) variation. We define the difference only in terms of angles since, as pointed out in previous works~\cite{guzov2021human}, the total distance tracked by the IMUs is well measured, while the curvature tends to accumulate drift over time. 

We then correct the human trajectory by optimizing the following energy: 
\vspace{-2mm}
\begin{equation}
F_{tr}(\myvec{l},\myvec{p}) = \underset{\hat{\alpha}}{\mathrm{arg\,min}} \left(
 \sum_{i=1}^{K}({||\hat{\myvec{l}}({\tau}_i)-{\myvec{p}}_i||_2}) + \lambda E_{tr}(\hat{\myvec{l}}, \myvec{l}) \right),
 \label{eq:optim}
\end{equation}
where $\lambda$ is the global rigidness coefficient, which encodes how much local angles should retain the initial estimation. 

\PAR{Contact-based human trajectory correction.} In \ours{}, we perform the above optimization two times. We consider the input trajectory recovered by the IMUs, and we optimize it using the control points returned by the camera localization. Then, our method detects the moments of contact along the human motion sequence. For each detection, we select the nearest object in the scene within a reasonable range (\eg, $\qty{50}{\centi\meter}$). The contact is ignored as a false positive if no object is that close. We select a contact point ${\myvec{p}}_c$ as the closest point of the object to the contacting hand.
Then, we rerun our optimization again, considering ${\myvec{p}}_c$ as the only control point to satisfy. We report details in supplementary.

\subsection{Contact detection} 
\label{sec:contact}
\PAR{Problem.} The key ingredient for accurate localization is detecting when and where the user interacts with an object. In this work, we purely focus on human poses (obtained from IMUs) for contacts instead of relying on camera data for multiple reasons: (1) contacts are often not visible, \ie, camera data alone is insufficient for the task. (2) IMUs are cheaper and much more power-efficient than cameras. At the same time, processing their lower-dimensional output requires significantly less compute (and thus power). This makes purely IMU-based contact prediction very attractive for applications running on mobile devices such as AR/VR headsets or robots. Naturally, combining inertial and visual data should improve performance, similar to visual-inertial localization. However, we leave this integration for future work and focus on IMU-only contact detection. 

\PAR{Training data.} Existing datasets for human-object contact prediction contain only a limited number of samples per object type~\cite{bhatnagar2022behave}, or only hand-held objects~\cite{taheri2020grab}. In our context, the interaction involves large objects appearing in real scenes. 
Hence, we collect and annotate a training dataset (H-contact, Sec.~\ref{exp:datasets}) of {\raise.17ex\hbox{$\scriptstyle\sim$}}680k pose frames ($>3$ hours) recorded with 8 subjects wearing IMUs and 12 different objects. Our dataset is noticeably bigger compared to several other human--object and human--scene interaction datasets (BEHAVE~\cite{bhatnagar2022behave} contains {\raise.17ex\hbox{$\scriptstyle\sim$}}15k frames, PROX~\cite{PROX:2019} {\raise.17ex\hbox{$\scriptstyle\sim$}}100k). 

\begin{figure*}[t]
    \centering
    \vspace{-0.3cm}
    \includegraphics[width=0.95\linewidth]{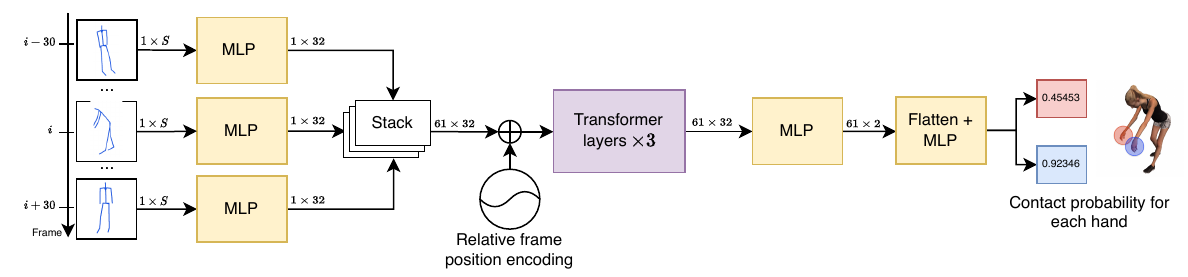}
    \vspace{-0.2cm}
    \caption{\label{fig:architecture}\textbf{Contact prediction based on human pose.} Interactions are frequently unobserved in an egocentric view, see Fig. \ref{fig:challenges} (top row), making contact prediction ill-posed. Instead, we propose to predict from sequences of full 3D human poses. We leverage a transformer-based architecture that takes 61 frames \{$i-30, \dots , i+30$\} of SMPL pose vectors of size $S=69$ and predicts the contact probability for each hand for the middle frame $i$. See Sec. \ref{sec:contact} for details.}

\vspace{-0.5cm}
\end{figure*}

\PAR{Transformer.} To predict contacts, we train a sequence-to-sequence Transformer~\cite{vaswani2017attention} to map a sequence of poses to a sequence of per-hand contact probabilities. 
Specifically, we concatenate $61$ SMPL pose vectors in a sequence, forwarding them to an MLP, appending the frame position as positional encoding, and processing them with a Transformer to output a sequence of contact probabilities for each hand. We use a sliding-window approach, and at each instance, we retain only the central (30\textsuperscript{th}) prediction. The contact is considered active once the probability reaches a certain threshold.
The architecture is visualized in Fig.~\ref{fig:architecture}.
To remove false negatives, any gap of $\leq \qty{0.5}{\second}$ between two active contacts is filled (\ie marked active). This produced the best results on a validation set (see supplementary).

While focusing on hands is not entirely descriptive of how humans interact with the world, it covers most cases in which humans cause changes in their environments. Our method can easily extend to other body parts; more detailed analyses are left for future works.

\PAR{Contact intervals.} Each group of consecutive frames with active contact is considered a \emph{contact interval}.
If the network predicts the end for one hand while the other is still considered to be in contact, \ours{} splits the contact interval into two interactions (a two-handed and a one-handed one). 
Similar cases (\eg, interchanging hands) are treated the same way. Likewise, our method can handle multi-object interaction -- please see supplementary for details.

\PAR{Training details.} 
We train the network for 100 epochs with a batch size of 100 using the Adam optimizer~\cite{DBLP:journals/corr/KingmaB14} with a learning rate of $10^{-3}$ and a 
binary cross-entropy loss. The resulting architecture has 21.9k parameters and an inference time of less than a second per minute of motion (3600 motion windows) on an Nvidia RTX 3090 GPU\@.

\subsection{Interaction modeling} 
\label{sec:modeling}

The benefits of \ours{}`s pose-based contact prediction and human localization are best visualized by dynamically adapting the scene changes as their consequence.
Concretely, when contact with an object is predicted, we attach the object to the user; its dynamic is driven by human motion given through IMU pose and the object's degrees of freedom defined in the interactive environment.
Please see the supplementary paper for the technical details.

\section{Experiments}
\label{sec:experiments}

\subsection{Datasets}
\label{exp:datasets}
In this work we captured and annotated two new datasets: \textbf{H-contact} and \textbf{EgoHOI}, which we release together with our annotation tool.

\PARnospace{H-contact} is a dataset of human--object interactions, designed to serve as a training set for our contact predictor. We captured and annotated more than 2300 human--object interactions in $>3$ hours of recordings divided into $30$ uninterrupted sequences. A total of around $680$k frames, providing interaction for $8$ subjects and $12$ objects, whose lengths range from $1$ to $19$ seconds.
To obtain ground-truth contact labels, we built a
GUI-based annotation tool for this task. Using synchronized video from an external camera, we asked annotators to define the contact classifications.

\PARnospace{Egocentric Human--Object Interactions (EgoHOI)} 
is a dataset of humans performing everyday interactions with objects in real scenes recorded with wearable sensors. The sensors are placed directly on the human to allow for large recording volumes not restricted by external camera placement. 
The wearable setup consists of the IMU-based motion-capturing suit Xsens Awinda \cite{xsensawinda}, allowing us to obtain human pose sequences, and a head-mounted RGB camera for visual localization of the subject in the scene.
The dataset also includes the related interactive environments (IE): a 3D scans of the scene, segmented objects and their degrees of freedom.
EgoHOI contains interactions with 14 objects (tables, windows, doors, drawers, sofas, chairs, \etc) in multiple IE for a total of more than 100k motion frames. We also recorded RGBD data from an external multi-camera setup to measure reconstruction accuracy.

\begin{figure*}[t]
\vspace{-4mm}
    \centering
    \small
    \begin{overpic}[trim=0cm 0cm 0cm 0cm,clip, width=0.94\linewidth]{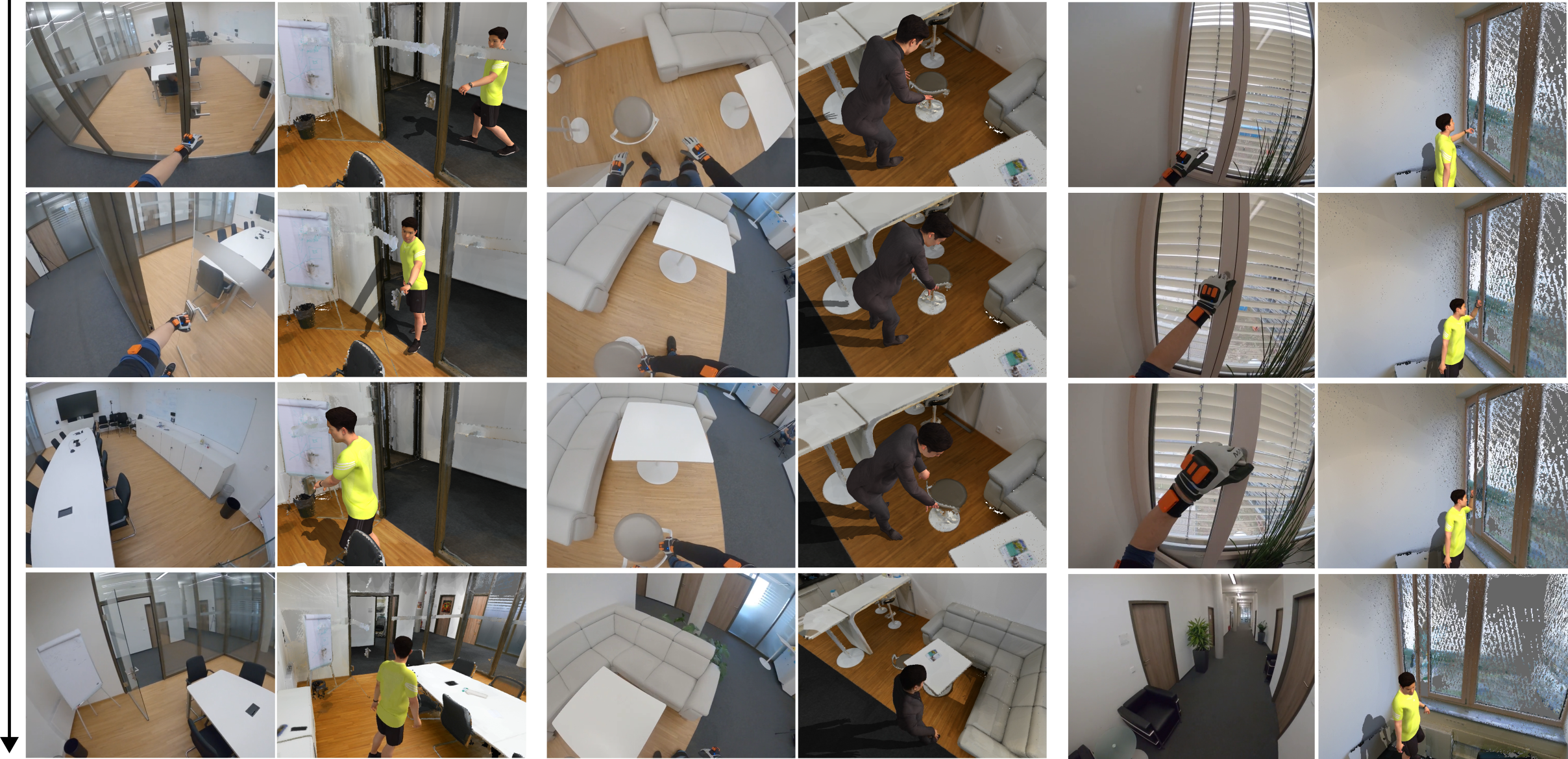}
    \put(-3,48){time }
    \put(13,49){Interaction 1 }
    \put(46,49){Interaction 2 }
    \put(79,49){Interaction 3 }

    \put(2.5,-2){Head camera view}
    \put(36,-2){Head camera view}
    \put(69.5,-2){Head camera view}

    \put(23,-2){\ours{}}
    \put(56,-2){\ours{}}
    \put(89,-2){\ours{}}
    
	\end{overpic}

    \vspace{0.3cm}
    \caption{\textbf{Qualitative results.} We show three examples of human interaction, pairing the head-mounted camera view with the interaction modeling achieved by iReplica. The object is not always visible during the interaction (Interaction 1), hand grasping can be difficult to understand from the camera (Interaction 2), or object occludes a majority of the first person view (Interaction 3). By relying on human-centric contact detection, iReplica achieves reliable modeling in all these challenging scenarios. Please see our video for more results.}
    \label{fig:qualitative_results}
    \vspace{-4mm}
\end{figure*}

\begin{figure*}[t]
    \centering
    \includegraphics[width=0.94\linewidth]{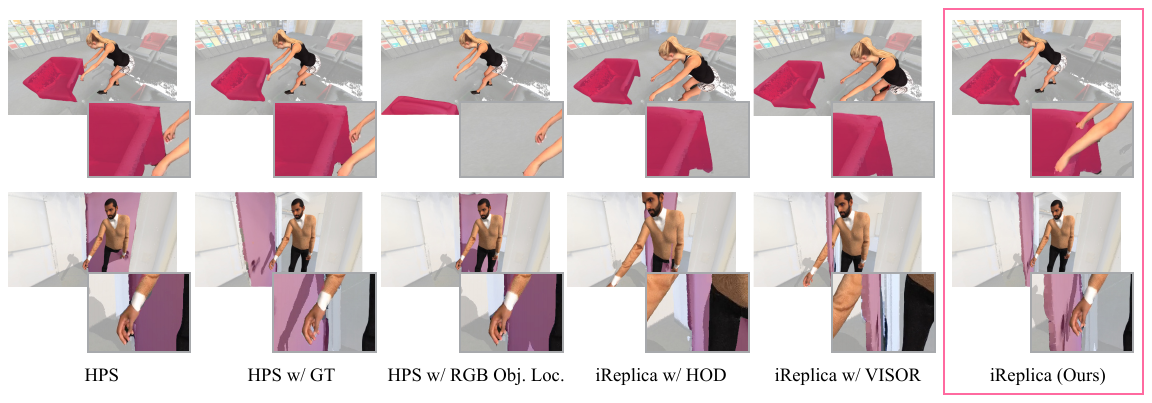}

    \vspace{-0.2cm}
    \caption{\textbf{Qualitative comparison.} We compare \ours{} (ours) to the baseline methods (interacted object highlighted in red for visual clarity). For the sofa sequence (top row) no baseline can track the sofa and correctly place the subjects` hands. Similarly, the door (bottom row) is incorrectly placed by all baselines, and the hand is not in contact with the handle. In contrast, \ours{} obtains visually plausible results by adjusting human and object locations to satisfy contact constraints.
    } 
    \label{fig:qualitative_comparison}

\vspace{-0.5cm}
\end{figure*}

\subsection{Baselines}
Due to the novelty of the proposed human--object tracking task, no published baselines exist.
We introduce novel baselines and briefly describe them, see supp.mat. for details. 

\PAR{HPS.}
We compare to HPS~\cite{guzov2021human} that localizes the human within the prescanned scene using the images of the head-mounted camera.
HPS does not reason about human--object interactions and does not track scene changes. 

\PARnospace{HPS w/ GT}  
combines HPS with ground-truth data to  
predict the object's motion. 
It has 
access to the ground-truth final object pose and ground-truth start and end time of the object interaction.
To obtain the object motion estimate, it linearly interpolates the object poses in the time window.
Obviously, the required ground-truth data for HPS w/ GT is not available in real-world applications.  
This baseline is used to show that a simple linear interpolation model is not sufficient for real-world scenes.

\PAR{HPS w/ RGB Obj. Loc.} localizes the object using solely the RGB frame from the head-mounted camera. As HPS uses a visual localization algorithm~\cite{sarlin2019coarse} to localize the camera in the scene, we use the same process to localize the interacted object w.r.t. the camera. Knowing the relative pose of the object to the camera localization in world space, we can estimate the object pose in world space. To simplify the matching process, this baseline uses GT object segmentations of the objects of interest.

\PAR{\ours{} w/ HOD~\cite{shan2020understanding} and \ours{} w/ VISOR~\cite{VISOR2022}.} 
Our approach predicts human-scene contacts  
based on human pose information.
Alternatively,  
the head-mounted RGB camera can be used  
to make these predictions.  
As baselines, we use two state-of-the-art, pretrained, RGB-based hand-contact understanding methods: \textbf{HOD}~\cite{shan2020understanding} and \textbf{VISOR}~\cite{VISOR2022}; both 
predict 2D hand and object locations, and contact probabilities per hand.
We use the contact probabilities as a drop-in replacement for the
contact predictor in \ours{}, while keeping the rest of the method fixed.

\subsection{Results}

\vspace{-0.2cm}

\PAR{Qualitative results.} 
Fig.~\ref{fig:qualitative_results} shows our results for sample frames from multiple-scene interactions.  
Videos are included in the supp.\@ mat.\@ and we urge the reader to look at them as interactions are best judged in motion. 
Our results show that egocentric motion data alone can localize the human in the scene, model the interaction between the human and objects, and update the scene accordingly. 
Our contact predictor allows \ours{} to estimate object tracking only based on the human pose;
\eg, doors, chairs, and tables can be interacted with in the scene.

\PAR{Qualitative comparisons to baselines.}
Fig.~\ref{fig:qualitative_comparison} visually compares \ours{} to our baselines by showing individual frames from some of the interactions. 
The interactions are best viewed in the supp.\@ video. 
\textit{HPS} does not track scene changes and thus obtains unrealistic motions. For example, the door opening is not tracked. The subject should have opened the door with the handle, but the door is still closed. (Fig.~\ref{fig:qualitative_comparison}, door). 
Or the sofa is dragged by the subject, but the object stands still. The sofa, therefore, is visibly not in contact with the subject's hands (Fig.~\ref{fig:qualitative_comparison}, sofa). 
\textit{HPS w/ GT} linearly interpolates the object motion given ground-truth object start and end pose and contact times. 
The resulting interaction is not visually plausible due to a large mismatch between the subject's hands and the sofa. 
Human-scene interaction motion is highly non-linear, so linear approximations seem unrealistic: according to our user study, iReplica results were preferred over this baseline and HPS in 84.2\% of the cases (see suppl.).
\textit{HPS w/ Obj.\@ Loc.} fails to detect the accurate object position during the interaction, resulting in misaligned results, to the point that it fails to localize the object at all (\eg, Tab.~\ref{table:contact_acc}, Box).
\textit{\ours{} w/ HOD} and \textit{\ours{} w/ VISOR} both suffer from false negatives, resulting in a sudden contact loss in the middle of the interaction and missed contacts between object and subject.
\textit{\ours{}} ensures that the subject's hands are close to the object during the whole interaction -- a key aspect of visual plausibility not achieved by the baselines.  
This shows the value of \ours{}'s correction of the human trajectory based on the human--object interaction.

\PAR{Reconstruction quality compared to real scenes.}
We quantitatively validate \ours{}'s object and human localization results in terms of the reconstruction quality with respect to the original scene. We measure deviations from the virtual replica to the real scene using the EgoHOI dataset.
Tab.~\ref{table:obj_loc_acc} shows the object localization accuracy  
at the end of the interaction, where the GT object pose was annotated.
On average, \ours{} improves the results considerably (col.~\textit{All}). All object types are localized with a distance below 10 cm and an orientation error  
below 13 degrees.

\PAR{Ablation of Contact-based human trajectory correction.}
We ablate the contact-based human trajectory correction by excluding it from \ours{}. We report the results in Table \ref{table:obj_loc_acc} and \ref{table:contact_acc}  (\textbf{iReplica w/o Contact corr.}). The method greatly benefits from the proposed correction.
We measure the error of human localization with (\ours{}) and without  (HPS~\cite{guzov2021human}) correction on a special sequence that additionally has ground truth point clouds obtained via an external multi-view system of depth cameras. \ours{} again improves upon HPS -- see the supp. mat. for details.

\PAR{Visual plausibility.} To measure the visual plausibility of \ours{} results compared to the baselines, we consider the contact between the human and the object. In particular, we measure the mean distance from the object to the interacting hand, see Tab.~\ref{table:contact_acc}.
\ours{} keeps this distance below 3 cm. 
Tracking contacts and using them for attaching the object to the human motion creates the lowest distances.

\PAR{Contact prediction accuracy.}
We benchmark the accuracy of \ours{} contact prediction in isolation in Tab.~\ref{table:net_metrics}, comparing it to our two RGB contact prediction baselines, HOD~\cite{shan2020understanding} and VISOR~\cite{VISOR2022}. We treat the network predictions as probabilities in a binary classification task and compute 4 metrics: Average Precision (AP), Precision, Recall and Accuracy on the binarization threshold of $0.5$. 
Our contact prediction, solely based on the 3D human pose, significantly outperforms the RGB-based reasoning - one cause is that interaction is not always visible in the camera. Once more, we remark on how 3D human poses in isolation is a highly informative indicator of interaction contacts.

\begin{table}
\vspace{-0.5cm}
\begin{center}
\scriptsize
	\tabcolsep=0.11cm
\begin{tabular}{llccccc}
\toprule[1.5pt]

Error  $\downarrow$ & \multicolumn{1}{l}{Method} & Door & Sofa & Table & Box & All \\
\midrule
\multirow{3}{*}{\begin{tabular}{@{}l@{}}Distance\\(in cm)\end{tabular}} & \textbf{HPS} & 79.27 & 69.54 & 25.31 & 41.92 & 60.81 \\
& \textbf{HPS w/ RGB Obj.\@ Loc.} & 28.66 & 1684.06 & 119.59 & --- & 597.83 \\  
& \textbf{\ours{} w/ HOD~\cite{shan2020understanding}} & 57.50 & 55.78 & 1.62 & \textbf{3.33} & 38.58 \\
& \textbf{\ours{} w/ VISOR~\cite{VISOR2022}} & 43.40 & 66.74 & 5.98 & 11.31 & 39.59 \\
& \textbf{\ours{} w/o Contact corr.} & 18.54 & 11.70 & 1.84 & 7.79 & 11.68 \\
& \textbf{\ours{} (Ours)} & \textbf{9.97} & \textbf{6.66} & \textbf{0.90} & 7.09 & \textbf{6.88} \\
\midrule
\multirow{3}{*}{\begin{tabular}{@{}l@{}}Angle\\(in \textdegree)\end{tabular}} & \textbf{HPS} & 109.19 & 23.53 & 12.16 & 3.76 & 46.89 \\
& \textbf{HPS w/ RGB Obj.\@ Loc.} & 34.36 & 118.02 & 60.08 & --- & 61.43 \\  
& \textbf{\ours{} w/ HOD~\cite{shan2020understanding}} & 75.74 & 7.74 & 0.78 & \textbf{2.71} & 28.41 \\
& \textbf{\ours{} w/ VISOR~\cite{VISOR2022}} & 56.64 & 17.36 & 2.87 & 12.78 & 27.27 \\
& \textbf{\ours{} w/o Contact corr.} & 22.16 & \textbf{5.83} & 0.88 & 4.81 & 10.28 \\
& \textbf{\ours{} (Ours)} & \textbf{12.94} & \textbf{5.83} & \textbf{0.43} & 4.81 & \textbf{7.13} \\

\bottomrule[1.5pt]
\end{tabular}

\caption{
\textbf{Object localization accuracy.} Distance (in cm) and angle (in \textdegree) between object center at the end of the interaction in the GT pose and object center in the  pose predicted by the algorithm.}
\vspace{-0.13cm}
\label{table:obj_loc_acc}
\end{center}

\end{table}

\begin{table}
\vspace{-0.5cm}
\begin{center}
\footnotesize
	\tabcolsep=0.11cm
\begin{tabular}{lccccc}
\toprule[1.5pt]

\multicolumn{1}{l}{Class label} & Door & Sofa & Table & Box & All \\
\midrule
\textbf{HPS} & 46.00 & 38.32 & 26.35 & 6.64 & 33.61 \\
\textbf{HPS w/ GT} & 17.28 & 6.90 & 7.55 & 6.74 & 10.44 \\
\textbf{HPS w/ RGB Obj.\@ Loc.} & 65.26 & 724.63 & 136.27 & --- & 287.12 \\
\textbf{\ours{} w/ HOD~\cite{shan2020understanding}} & 48.42 & 35.96 & 13.31 & 5.52 & 31.26 \\
\textbf{\ours{} w/ VISOR~\cite{VISOR2022}} & 33.76 & 51.14 & 13.39 & \textbf{3.84} & 31.17 \\
\textbf{\ours{} w/o Contact corr.}  & 18.15 & 9.80 & 6.89 & 5.45 & 11.37 \\
\textbf{\ours{} (Ours)} & \textbf{2.83} & \textbf{1.46} & \textbf{3.49} & 5.49 & \textbf{2.93} \\
\bottomrule[1.5pt]
\end{tabular}
\caption{
\textbf{Visual plausibility of human-scene interaction.} Mean distance between the object and the contacting hand (in cm) over the interaction time.}
\label{table:contact_acc}
\end{center}

\end{table}

\begin{table}
\vspace{-5mm}
\begin{center}
\scriptsize
	\tabcolsep=0.11cm
\begin{tabular}{lcccc}
\toprule[1.5pt]
\multicolumn{1}{l}{Contact predictor} & AP $\uparrow$ & Precision@0.5 $\uparrow$ & Recall@0.5 $\uparrow$ & Accuracy@0.5 $\uparrow$ \\
\midrule

\textbf{HOD~\cite{shan2020understanding}  
} &  0.044 & 0.251 & 0.818 & 0.364 \\
\textbf{VISOR~\cite{VISOR2022} 
} & 0.217 & 0.313 & 0.098 & 0.732 \\
\textbf{Ours  
} & \bf0.807 & \bf0.786 & \bf0.880 & \bf0.905 \\

\bottomrule[1.5pt]
\end{tabular}
\caption{
\textbf{Contact prediction performance.} Metrics obtained on our test set with subjects that are not appearing in training data.
}
\label{table:net_metrics}
\end{center}
\vspace{-3mm}
\end{table}

\section{Discussion and Conclusion}
\label{sec:conclusions}

We proposed the novel problem of capturing human-scene interactions and dynamic 3D scenes solely from wearable sensors - that is, IMUs and a head-mounted camera, and not relying on any external cameras or object trackers.
We show that egocentric data alone can be used to localize the human in the scene, model the interaction with the objects, and update the scene accordingly. \ours{} enhances results of human localization from HPS, correcting that the human and the interacted object are close to each other.

\PAR{Future work and Limitations.} Our simple method has some natural limitations, which point to interesting future directions. Our approach does not consider physics or collisions, which future work can investigate using simulations similar to those available in game engines. Our work relies on a prescanned IE, in which objects degrees of freedom are known a priori. Incorporating environment reconstruction (e.g., SLAM-based models) and on-the-fly segmentation (e.g., ScanNet) could remove these assumptions. Finally, our paradigm does not consider more complex interactions and object manipulations (e.g., articulated, non-rigid). This would be an exciting research direction, especially in light of the recent availability of human-object datasets~\cite{fan2023arctic}.

{\small
\PAR{Acknowledgments:} Special thanks to RVH team members, and reviewers, their feedback helped improve the manuscript. The project was made possible by funding from the Carl Zeiss Foundation. This  work is supported by the Deutsche Forschungsgemeinschaft (DFG, German Research Foundation) - 409792180 (Emmy Noether Programme, project:  Real Virtual Humans), German Federal Ministry of Education and Research (BMBF): Tübingen AI Center, FKZ: 01IS18039A and the Czech Science Foundation (GA\v{C}R) EXPRO (grant no. 23-07973X). Gerard Pons-Moll is a member of the Machine Learning Cluster of Excellence, EXC number 2064/1 - Project number 390727645. Julian Chibane is a fellow of the Meta Research PhD Fellowship Program - area: AR/VR Human Understanding. Riccardo Marin has been supported by the European Union’s Horizon 2020 research and innovation program under the Marie Skłodowska-Curie grant agreement No 101109330. 
}

\clearpage

{\small
\bibliographystyle{ieeenat_fullname}
\bibliography{egbib, torsten}
}

\clearpage

\vspace{-9mm}
	
\twocolumn[{%
\renewcommand\twocolumn[1][]{#1}%
\newpage
\null
\vskip .375in
\begin{center}
  {\Large \bf SUPPLEMENTARY MATERIALS \\ Interaction Replica: Tracking human--object interaction and scene changes from human motion \par}
  \vspace*{24pt}
\end{center}
}]

\begin{abstract}
    In this document, we report additional details and results, which we cannot include in the main manuscript for space limits: we present a general description of HPS summarizing its key aspects relevant to our method (Sec.~\ref{sec:hps}); we report the details of our capturing setup (Sec.~\ref{sec:capturing}) and how we can incorporate data coming from motion capture gloves (Sec.~\ref{sec:hands}); we detail our framework for Interaction Modeling: how we validate our false-positive pruning, model objects motion driven by human contact, and our strategy to deal with multiple objects (Sec.~\ref{sec:modelingsupmat}); we show an intuition behind the trajectory correction and the regularization of the penalty for angles deviations (Sec.~\ref{subsec:meth_control}); we provide information about implementation and performance (Sec.~\ref{sec:performance}) and detailed descriptions of baselines (Sec.~\ref{sec:baselines}); additional comparison for the contact-based prediction model (Sec.~\ref{sec:contact_comparisons}); an external RGBD camera evaluation which further prove the relevance of human-object interaction also for human localization (Sec.~\ref{sec:external}); examples of our annotation tool (Sec.~\ref{sec:annotation}) and of the dataset (Sec.~\ref{sec:examples}) and \ours{}'s features (Sec.~\ref{sec:moreres}). We hope that this extensive collection of resources can help the reader's intuition and ease the work on this complex problem for the future researchers.
\end{abstract}

\section{Human POSEitioning System (HPS)}
\label{sec:hps}
For completeness, we briefly introduce HPS, the system we use in our pipeline to provide initial human poses.

The HPS pipeline produces poses for the SMPL~\cite{smpl2015loper} model together with the global position within the 3D scan coordinate system. It uses data from the body-mounted sensors, the video from the head-mounted camera, and the 3D point cloud of the pre-scanned scene as input. The output represents the human localized in the 3D scene. HPS relies on three steps:

\textbf{1) Pose estimation from XSens.}
From the IMU measurements, XSens uses a proprietary algorithm based on the Kalman filter and a kinematic body model, which takes physical limitations of the body into account. The system provides the human pose and translation estimation in its own coordinate frame. While these estimates can be considered accurate, the system has two main drawbacks: it accumulates errors over time, leading to significant drift, and the human dynamic is not registered within the global 3D scene, causing potential inconsistencies (\eg, wall and floor penetration).

\textbf{2) Head-mounted camera localization.} 
For every camera frame, HPS provides the initial head pose estimates using a hierarchical localization algorithm~\cite{sarlin2019coarse}, which operates on RGB images. Given a dataset of images with known positions in the pre-scanned scene, the algorithm establishes correspondences between those images and the camera frame, finding the position of the head camera by minimizing the reprojection loss. The camera lets HPS globally localize the human in the scene, but the raw localizations are noisy and do not ensure realistic human placement in the scene.

\textbf{3) Combining IMUs and the head-mounted camera.}
HPS combines the two previous steps together with the 3D point cloud of the pre-scanned scene to optimize the human pose and satisfy all the previously mentioned constraints.
The simultaneous optimization for the IMU-based body poses and camera localizations eliminates the noise of the camera pose predictions, and significantly decreases IMU drift at the same time, resulting in stable and accurate results.

\section{Capturing Setup}
\label{sec:capturing}
We use a combination of the Xsens Awinda~\cite{xsensawinda} IMU system to capture body motions and a head-mounted camera (Apple iPhone 12 or GoPro Hero 8) to capture the first-person view.
The camera captures at a resolution of $1920 \times 1440$, $30$ FPS. While both cameras feature automatic stabilization, we explicitly turn it off to get a better idea of the head motion. We use OpenCV~\cite{opencv_library} to calibrate the camera intrinsics.  
The Xsens Awinda consists of 17 IMU sensors attached to the body with velcro straps and captures the body motion at $60$ FPS. 

The Xsens Awinda outputs human poses in the form of skeleton joint angles, which is not directly compatible with the SMPL~\cite{smpl2015loper} pose parameters format. To convert between these formats, we developed the following retargeting algorithm: we export the motion from the Xsens internal format (MVN) into the Autodesk FBX format and use the Autodesk FBX Python SDK~\cite{autodesk_fbx} to extract the rotation of each joint as a quaternion. We convert those quaternions to the axis-angle representation used in SMPL. We map joint rotations from the FBX skeleton to the SMPL skeleton according to a manually designed mapping. Finally, we produce the SMPL pose parameter vector by concatenating the mapped axis-angle joint rotations in the right order. If available, we additionally use Xsens gloves from Manus to capture the fine-grained fingers motion, and adapt the algorithm accordingly (see Sec.~\ref{sec:hands}).

\section{Adding Hands Data}
\label{sec:hands}
If fine-grained hand positions are available (\eg, captured with motion capture gloves), we can modify the algorithm to consider such data. Namely, we replace the SMPL model with SMPL+H~\cite{MANO:SIGGRAPHASIA:2017}, which has the same template body mesh, but provides additional 30 joints (3 joints for each finger) for detailed hand pose representation. We additionally change the input of our contact prediction network to accept vectors of concatenated body pose and hand pose parameters, therefore the input becomes 61 vectors of size $1\times \hat{S}$, $\hat{S}=159$, adding 90 parameters of hands pose to each input vector. No additional architecture changes are made.

\section{Interaction Modeling - Details}
\label{sec:modelingsupmat}
\begin{figure}[H]
\begin{minipage}{\linewidth}
    \centering
    \includegraphics[width=\linewidth]{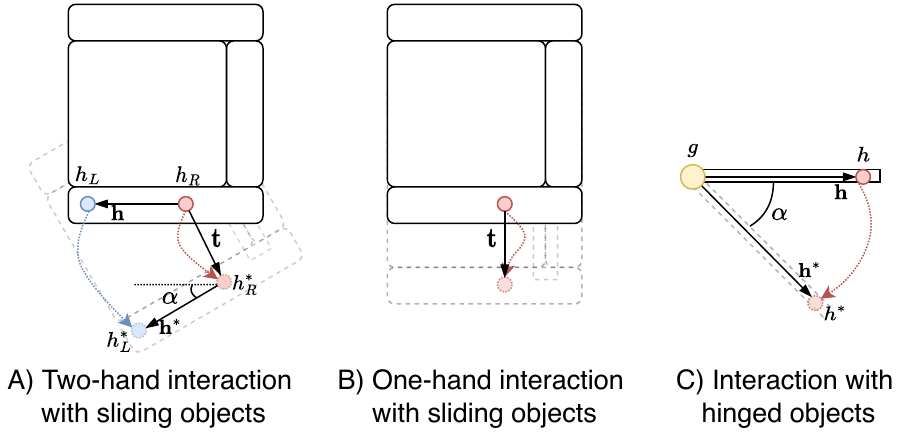}
    \caption{\label{fig:objtrajectory}\textbf{Obtaining object trajectory from hand interactions}. Colored dashed lines denote hands trajectories, $\alpha$ is inferred rotation angle, $\mathbf{t}$ is inferred object translation vector}

\end{minipage}
\end{figure}

In this section, we provide more details on the object motion driven by human contact, also depicted in Fig.~\ref{fig:objtrajectory}. 

\setlength{\columnsep}{2pt}
\begin{wraptable}[5]{R}{0.4\linewidth}
	\footnotesize
	\vspace{-0.8cm}
    
   \begin{center}	    	
	\hspace{-1cm}

	\begin{tabular}{lr}
               $t$               & Error (cm) \\ \hline
		$0.00$                 &  $13.61$  \\
		$0.25$              &  $8.49  $    \\
		$\mathbf{0.50}$      &  $\mathbf{7.13}$   \\ 
           $1.00$                 &  $8.86$   \\ 
           $2.00$                 &  $8.86$   \\ 

	\end{tabular}
   \end{center}
\end{wraptable}
\textbf{Processing the contact intervals.}
To remove false negatives, we post-process the predicted sequence and fill in gaps between active contacts that are shorter than $\qty{0.5}{\second}$, which produced the best results on a validation set (see the inset table).

\textbf{Degrees of freedom.} We model the degrees of freedom (DOF) of the motion of each object as to avoid unrealistic motions. For example, a door can only rotate around a specific axis, and a sofa can only slide along the floor.
While our model operates in a 3D scene, we present all derivations in 2D, since all the processed motions happen along the floor and any change in the direction orthogonal to the movement plane does not affect the object trajectory and will be removed.

\paragraph{A) Two-hands interaction with sliding objects.} We consider the point cloud of an interactive object $O \in \mathbb{R}^{n \times 3}$ that can be freely moved on a two-dimensional plane. When the two hands contact the object, we denote the position of two keypoints corresponding to the middle of each hand as $h_L = (x_L, y_L)$ and $h_R = (x_R, y_R)$ for the left and the right hand, respectively, and $\mathbf{h}$ as the vector that connects $h_R$ to $h_L$. When the human moves, we register the new positions of the hands as ${h}_L^*$ and ${h}_R^*$, together with the new connecting vector $\mathbf{h}^*$. Then, we compare the hand configurations to recover a translation and a rotation. Firstly, we define the translation $t$ as
\begin{equation}
\mathbf{t} = ({x}_R^* - x_R, {y}_R^* - y_R).
\end{equation}
Then, we compute the rotation angle as
\begin{equation}
\alpha = \arccos\left(\frac{\mathbf{h}\mathbf{h}^*}{|\mathbf{h}| |\mathbf{h}^*|}\right).
\label{eq:rot}
\end{equation}
Finally, we recover the 2D rotation matrix $\mathbf{R}^\alpha$ associated with the angle. The sign of $\alpha$ encodes the direction of the rotation, and it can be obtained by taking the cross-product between the vectors $\mathbf{h}$ and $\mathbf{h}^*$.

\paragraph{B) One-hand interaction with sliding objects.} When only one hand $h$ interacts with the object, without any further information about the object's physics (\eg, its friction with the ground), it is impossible to recover the object rotation. Hence, given a new configuration ${h}^*$, we just compute the translation
\begin{equation}
\mathbf{t}  = (\widehat{x} - x, \widehat{y} - y).
\end{equation}

\paragraph{C) Interaction with hinged objects.} In this case, the object has a hinge positioned in $g = (x_g, y_g)$. When the contact begins at the point $h = (x_h, y_h)$, we compute the vector $\mathbf{h}$ that connects $g$ to $h$:
\begin{equation}
    \mathbf{h}=(x_h - x_g, y_h - y_g).
\end{equation}
When the human moves, we register the new position of the contact point ${h}^*$ and accordingly recompute the connecting vector $\mathbf{h}^*$. Then we compute the angle $\alpha$ between $\mathbf{h}$ and $\mathbf{h}^*$ as in Equation~\ref{eq:rot}, and we recover the associated rotation matrix $\mathbf{R}^\alpha$.

\vspace{0.5cm}

After obtaining these transformations, we apply them to the first two coordinates of each point of the object $O$.

\begin{equation}
    O^* = \mathbf{R}^\alpha O + t
\end{equation}

\subsection{Multiple Objects Interaction}
The contact strategy described in Sec.~\refg{4.2} of the main paper naturally extends to multiple objects interaction. When the start of contact is identified, \ours{} considers the closest object within a 0.5 m radius (if any) and initiates the interaction, which lasts for the whole contact interval. After the end of the contact, the object is released, and the method waits for the next contact interval to proceed with the next interaction.

To better distinguish between the objects and improve our robustness to false positive interactions, the contact predictor of \ours{} comprises a set of transformers, one for each interaction class: one for the sliding objects and one for the hinged ones. These two networks work in parallel: when one detects a starting contact, the object search in the neighbourhood is performed only for the specific category. The contact is ignored if no object of that class is identified in the user's proximity.

\section{Trajectory Optimization - Details}
\label{subsec:meth_control}
In Equation~\refg{1} of the main manuscript, we introduced the $E_{tr}$ energy, which limits the deviation of the trajectory, penalizing changes in the angles. Such energy is weighted by a coefficient $\lambda$ in eq.~\refg{2} of the main manuscript, which defines a global rigid coefficient of the trajectory: a lower value lets the trajectory freely bent to fit the control points, while a higher value preserves the local curvature and promotes global rigidness of the trajectory. To help the reader's intuition, we report in Fig.~\ref{fig:meth_physics_simulation} an example for different values of $\lambda$.

For completeness, we also report the definition of $\text{atan2}$ function:

\begin{figure}[t]
    \centering
    \includegraphics[width=\linewidth]{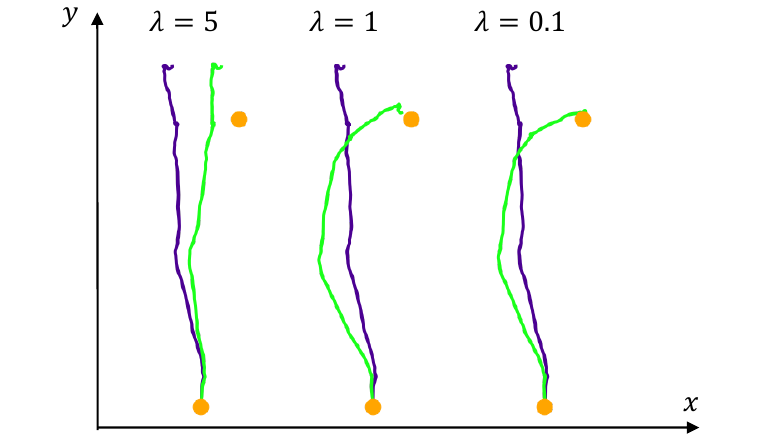}
        \caption{\textbf{Trajectory fitting with bending energy.} Bending trajectories with $F_{tr}$ using different rigidness coefficients $\lambda$, purple marks the original trajectory, green marks the result, orange dots denote control points.}
        \label{fig:meth_physics_simulation}
\end{figure}

\begin{equation*}
\operatorname{atan2}(y,x)=\begin{cases}\arctan({\frac {y}{x}})&{\text{if }}x>0,\\\arctan({\frac {y}{x}})+\pi &{\text{if }}x<0{\text{ and }}y\geq 0,\\\arctan({\frac {y}{x}})-\pi &{\text{if }}x<0{\text{ and }}y<0,\\+{\frac {\pi }{2}}&{\text{if }}x=0{\text{ and }}y>0,\\-{\frac {\pi }{2}}&{\text{if }}x=0{\text{ and }}y<0,\\{\text{undefined}}&{\text{if }}x=0{\text{ and }}y=0.\end{cases}
\end{equation*}

\section{Implementation and Performance}
\label{sec:performance}
We implement our algorithms in Python using the PyTorch~\cite{NEURIPS2019_9015} library for the contact prediction network and the bending energy optimization algorithm. For the latter, we use the Adam~\cite{kingma2014adam} optimizer with $1000$ iterations, with the learning rate and the rigidness coefficient $\lambda$ acting as hyperparameters. 

The most computationally expensive part of our algorithm is the visual localization pipeline needed for HPS, requiring $8$ seconds per frame on an NVIDIA Q8000 GPU, while the other steps take little additional time. Such a computationally expensive pipeline provides good localization results and supports our exploration. Note that the performance of the visual localization network itself is not the focus of our study, and we expect that more efficient alternatives can replace this algorithm in the future.

\section{Baselines}
\label{sec:baselines}
In this section, we provide additional details for some of the baselines used in the paper.

\subsection{HPS w/ GT} In this baseline, we assume that an oracle provides the ground-truth final object position, as well as the time window of the interaction. We stress that neither of these is available at inference time in our setting, and our method does not rely on them. Then the human motion is solely estimated using HPS, and the object movement is modeled using linear interpolation for translation and spherical linear interpolation (Slerp~\cite{shoemake1985animating}) for rotation. 
By inspecting the qualitative results (please refer to the attached video), we see that motion between humans and objects happens asynchronously and, therefore, unrealistically. 
This baseline shows that, even in the presence of further assumptions, modelling the object trajectory is non-trivial. It is clear that human motion is rarely linear, and more sophisticated techniques are required.
\paragraph{User study.} To measure the realism of the motion produced by linear interpolation, we did a user study and asked 73 respondents to rank the interactions produced by \ours{}, HPS w/ GT and HPS by realism. Each participant was given 9 questions, each showing results generated by two methods side-by-side in a random sequence. \ours{} results were preferred in 84.2\% of the cases, proving that our body-driven object tracking produces more realistic interaction.

\subsection{HPS w/ RGB Obj.\@ Loc} In this baseline, we assume that for each frame of the head-mounted camera, we have a perfect 2D segmentation mask for the object, obtained by semi-automatic annotation using an interactive segmentation pipeline~\cite{sofiiuk2020f}. Then we use the same localization method~\cite{sarlin2019coarse} as HPS to provide a 6-DoF localization of the object \wrt the human. Starting from the dataset of images with the known object positions in the pre-scanned scene, the algorithm establishes correspondences between those images and frames from the head-mounted camera. The head-mounted camera is then localized by minimizing reprojection loss. Next, the object's location relative to the camera is recovered by matching and optimizing with only the 2D key points inside the object mask. This information is combined with the camera position to recover the object's location in world coordinates. 
This baseline highlights that the camera is unreliable for localizing the object in the space. Detecting local landmarks is dramatically harmed by occlusions, head shaking, and the object missing in several frames. 

\subsection{\ours{} w/ HOD} Given the availability of a head-mounted camera, we explore the possibility of using it to predict contacts. In this baseline, we replace our pose-based contact predictor with HOD~\cite{shan2020understanding}, a method trained on many YouTube videos. Starting from a single RGB image, it predicts a full set of hand interaction properties: hands bounding box, object bounding box, and the contact state for each hand. We keep the rest of our method fixed except for the contact prediction part. 

The original paper~\cite{shan2020understanding} shows several results from an egocentric perspective, but it also mentions failure cases when hands and objects are close to each other. We confirm this by qualitative inspection, noting that the method loses contact with the object during the interaction.

\subsection{\ours{} w/ VISOR} Given that HOD is trained on a variety of videos, which also include extrinsic views, we deploy a similar baseline but rely on an RGB method specifically trained on egocentric views. Specifically, we consider the baseline trained for the HOS challenge\footnote{https://github.com/epic-kitchens/VISOR-HOS} on the VISOR~\cite{VISOR2022} dataset. This baseline relies on PointRend~\cite{kirillov2020pointrend}, augmented with auxiliary detection heads to predict the contact, following the same schema of HOD~\cite{shan2020understanding}. As in the previous baseline, we replace our pose-based contact prediction, leaving other parts of the method untouched. However, also in this case, we observe missing contacts and wrong release prediction, often causing human penetration (\eg, crossing the door).

\begin{table}
\begin{center}
\scriptsize
	\tabcolsep=0.11cm
\begin{tabular}{lcccc}
\toprule[1.5pt]
\multicolumn{1}{l}{Contact predictor} & AP $\uparrow$ & Precision@0.5 $\uparrow$ & Recall@0.5 $\uparrow$ & Accuracy@0.5 $\uparrow$ \\
\midrule

\textbf{POSA~\cite{Hassan:CVPR:2021}%
} & 0.033 & 0.115 & 0.716 & 0.297 \\
\textbf{Ours %
} & \bf0.807 & \bf0.786 & \bf0.880 & \bf0.905 \\

\bottomrule[1.5pt]
\end{tabular}
\caption{
\textbf{Additional contact prediction comparison.} Additional comparison with the prediction model from POSA~\cite{Hassan:CVPR:2021}. Model was finetuned on our training data and tested on the same testing set as in Table~\refg{3} of the main paper.
}
\label{table:supp_net_metrics}
\end{center}
\end{table}

\section{Contact Prediction -- More Results}
\label{sec:contact_comparisons}
We present an additional comparison with a method of contact detection appearing in POSA~\cite{Hassan:CVPR:2021}. POSA presents a human-scene interaction model that can be used as a prior for human placement in the scene. Compared to iReplica, POSA has a different goal and applications: it does not consider dynamic interactions, works only with one human pose frame at a time and is focused on static scenes. However, this is the closest baseline for our task of temporal pose-based interaction prediction.
Results are presented in Table~\ref{table:supp_net_metrics}. For a fairer comparison, we finetune the POSA model on the same training subset from the H-Contact dataset used for iReplica; we also average contact prediction scores from a selected region of each hand and treat this as a per-hand contact probability.

The comparison shows that the POSA model could not estimate contacts reliably, even after finetuning it on our training data. One possible reason for this could be that this method works with single frames, which leads to prediction uncertainty for many poses, while iReplica's transformer-based model considers a one-second window of motion.

\begin{figure}
    \centering
    \includegraphics[width=\linewidth]{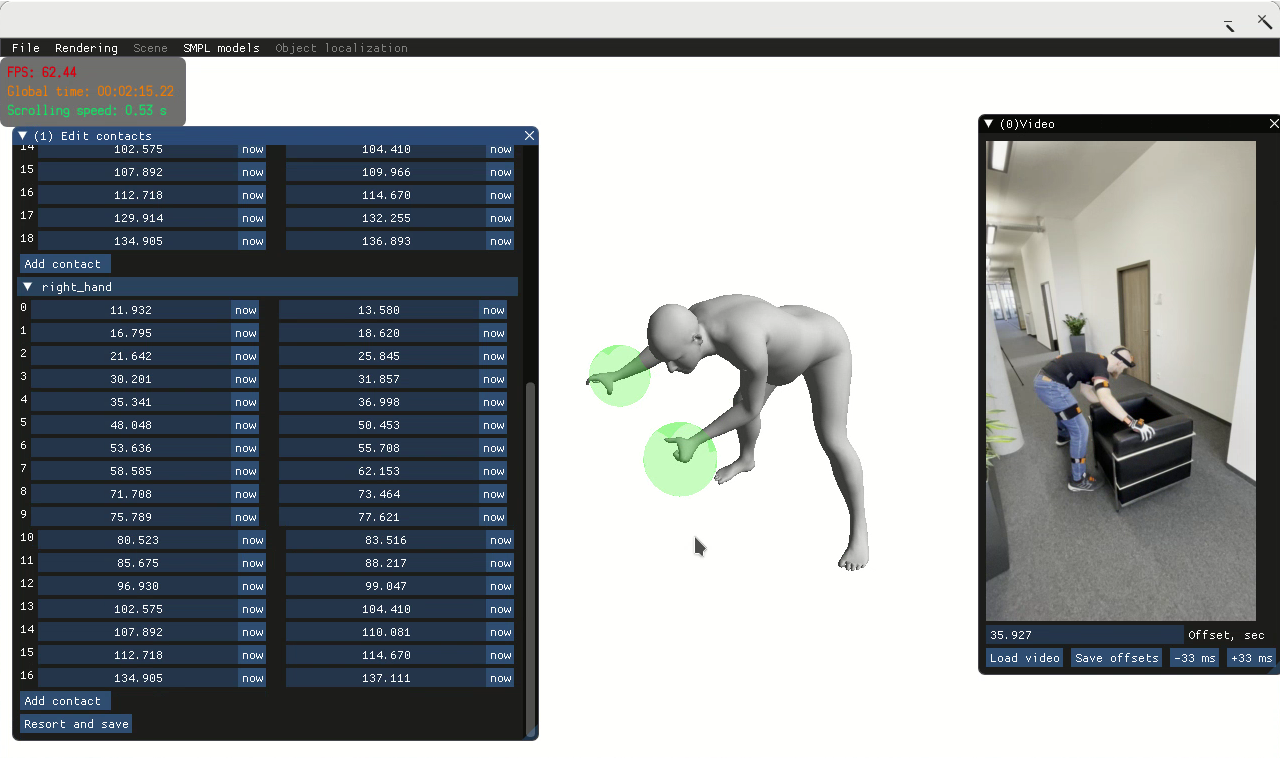}
        \caption{\textbf{Annotation tool.} The user views the RGB video frames (\emph{right}) and annotates the start and the end of contact interaction (\emph{left}). To help with disambiguating occlusions, the tool also shows the 3D pose (\emph{center}) together with the annotated presence of contact for each hand (green circles).}
        \label{fig:supp_ann_tool}
\end{figure} 

\begin{figure*}[t]
    \centering
    \includegraphics[width=\linewidth]{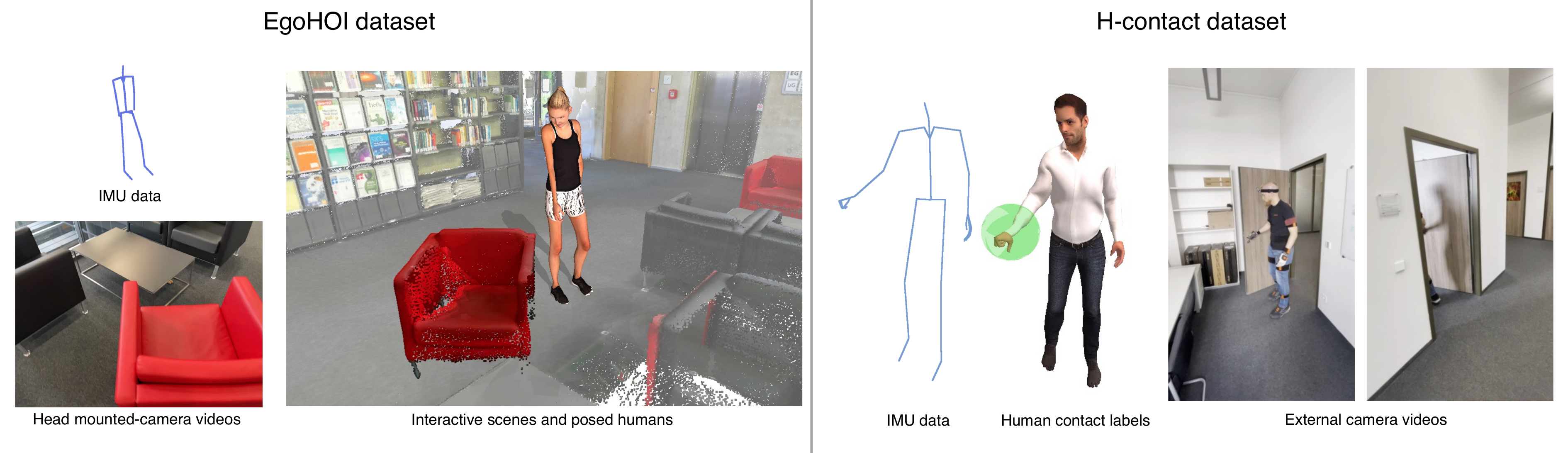}
        \caption{\textbf{EgoHOI and H-Contact examples.} For EgoHOI, we report for each timestamp the data obtained by the IMUs, the head-mounted camera frame, and the 3D posed human inside the interactive scene. For H-Contact, we provide IMUs data, contact labels for each hand, and recordings from external cameras.}
        \label{fig:supp_datasamp}
\end{figure*}

\section{External Camera Evaluation}
\label{sec:external}
\begin{table}

\begin{center}
\small
\tabcolsep=0.11cm
\begin{tabular}{lcc}
\toprule[1.5pt]
  & {Human to GT $E_{body}$}  $\downarrow$ & {Object to GT $E_{obj}$}  $\downarrow$\\
\midrule
\textbf{HPS}
 & 9.771  & 22.651 \\
\textbf{\ours{} (ours)} & \bf8.981 &  \bf8.471 \\
\bottomrule[1.5pt]
\end{tabular}

\caption{
\textbf{Human and object tracking quality \wrt the real scene:} Mean 3D error (in cm) between tracked and moving human/object models compared to the real scene. The scene is captured via a synchronized, multi-view RGBD video recording setup observing the interaction.}
\vspace{-0.13cm}
\label{table:external_cameras_results}
\end{center}
\end{table}

To additionally measure the human-object localization accuracy of our method, we recorded a special sequence with ground-truth data obtained via an external multi-view system of 3 depth cameras. The experimental setup closely follows the one used in HPS~\cite{guzov2021human}, however we capture the full dynamic object interaction while in HPS only the human body motion and the static scene was captured.

We use 3 calibrated Azure Kinect~\cite{azurekinect} RGBD sensors. By combining the outputs of these sensors, we obtain a sequence of 3D point clouds of the scene and a subject. Each sensor outputs the depth map with a resolution of $640 \times 576$ pixels and color frames with a resolution of $2048 \times 1536$ at around $30$ FPS. The Azure Kinect features built-in temporal synchronization, but to merge the output of the sensor into the scene ground truth representation, we also need to calibrate them spatially. For that we use a three-stage localization pipeline, similar to~\cite{guzov2021human}: \textbf{1)} we record a special sequence with 300 frames depicting the empty scene. We localize the RGB camera in each of these frames using the same algorithm as used for the head-mounted camera and average the 300 localization results; \textbf{2)} we perform ICP between the scene 3D scan and the point cloud unprojected from the depth map; \textbf{3)} we apply manual corrections if needed. Using the obtained positions of the sensors, the point cloud representation of the scene is formed by unprojecting depth maps from all 3 sensors to 3D. To perform the evaluation, we manually synchronize the time between the \ours{} motion sequence and the aforementioned point cloud representation.
For each frame of the test sequence, we separately measure object and human localization accuracy $E_{obj}$ and $E_{body}$:
\begin{itemize}
    \item $E_{obj}$: mean Chamfer distance from the object point cloud to the ground-truth point cloud,
    \item $E_{body}$: mean Chamfer distance from the human body SMPL mesh to ground-truth point cloud.
\end{itemize}

Results are presented in Table~\ref{table:external_cameras_results}.
Since HPS models only humans, the large error from the object ground truth is not surprising.
Although HPS and \ours{} share the same camera localization principle, we observe that our \ours{} improves human localization.
This validates that using the detected human--object interaction to adapt the human localization trajectory helps to improve reconstruction correctness. Moreover, it drastically enhances visual plausibility, as explained in the qualitative analysis.

\section{Annotation Tool}
\label{sec:annotation}

Fig.~\ref{fig:supp_ann_tool} is a screenshot of our annotation tool for preparing our H-contact dataset. Given a video frame (\emph{right}) and the reconstruction (\emph{center}), the user can annotate contacts by clicking on the interacting hands. The annotator can seek forwards and backward in the video, and the 3D reconstruction helps to disambiguate occluded poses. On average, it takes around 2--4 seconds to annotate 1 second of video.

\section{Examples from the Datasets}
\label{sec:examples}
In Fig.~\ref{fig:supp_datasamp} we report some examples from the H-contact and EgoHOI datasets. For H-contact, we provide IMU measurements, SMPL parameters, contact annotations, and external camera recordings. For EgoHOI, we provide interactive scenes, recordings from head-mounted RGB cameras, IMU data, as well as contact labels and GT final object positions (for evaluation purposes).

\section{Features of \ours{}}
\label{sec:moreres}
\begin{figure*}[t]
    \centering
    \includegraphics[width=\linewidth]{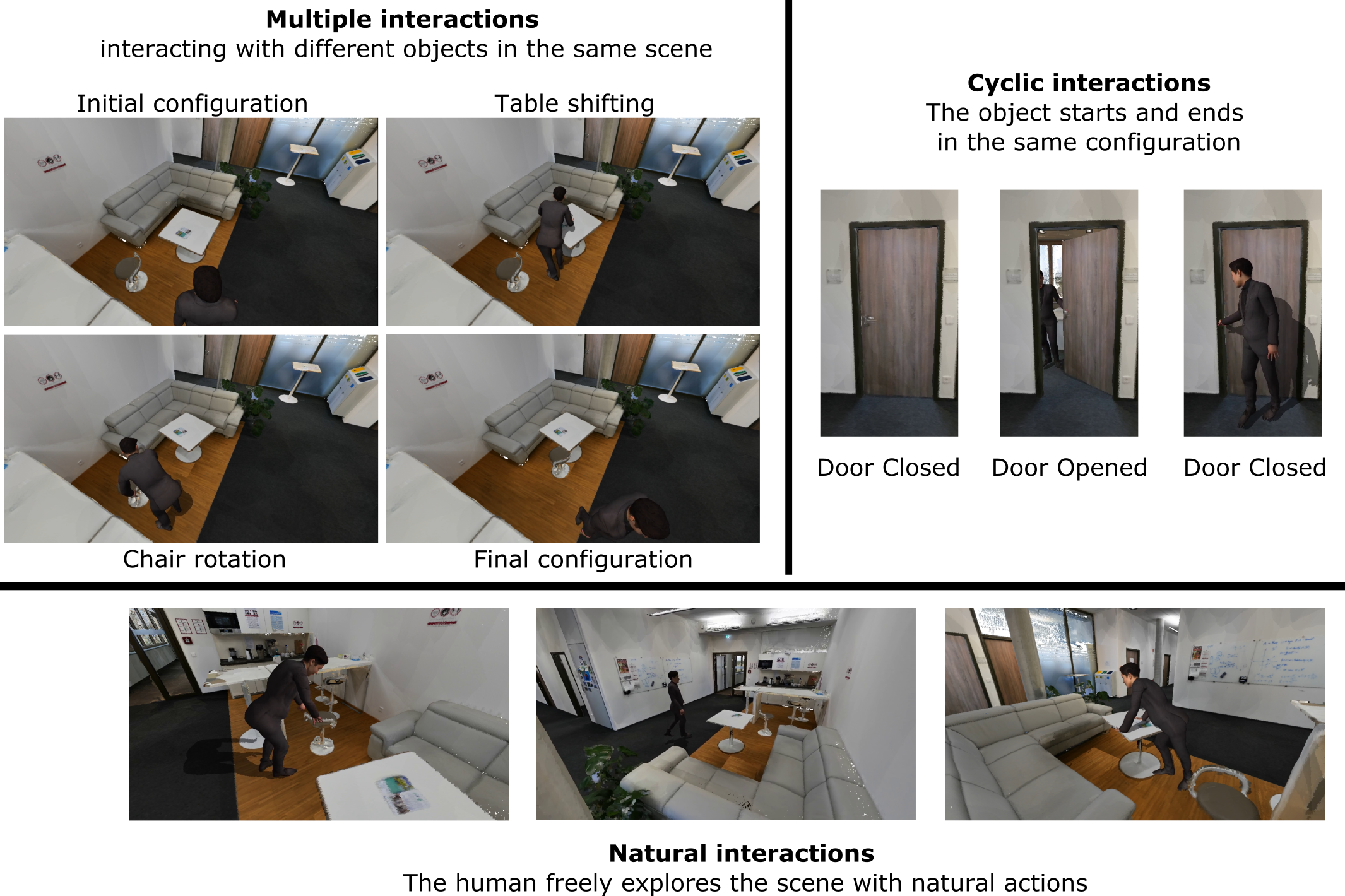}
        \caption{\textbf{Features of \ours{}.} \ours{} handles different kinds of challenges. It can be naturally applied to interactions involving several objects (\eg, arranging a table and a chair), objects that have a cyclic behavior in the scene (\eg, doors that open and close several times), and general daily interactions where the user freely moves in the space (\eg, walking to the kitchen, pushing a table). Please see the supplementary video for animated results.}
        \label{fig:hops-handles}
\end{figure*} 
Fig.~\ref{fig:hops-handles} highlight the features of \ours{} on the sequence, recorded to simulate a realistic and natural scenario. \ours{} does not assume a single object interaction: instead, it applies to interactive scenes where many objects can be differently re-arranged by the interaction (\eg, table, chair). \ours{} also works with objects with non-linear motion in space, for example, the doors that might start and end their movement in the same place. In this case, for example, an interpolation baseline would not provide any object dynamic since the initial and final configurations are the same. Instead, \ours{} tracks the full trajectory of the object. \ours{} allows the user to move in the space freely and interact naturally as in everyday life. We refer to the supplementary video for the animated results.

\clearpage

\end{document}